\documentclass[a4paper,UKenglish,cleveref, autoref, thm-restate]{lipics-v2021}

\hideLIPIcs  
\nolinenumbers

\title{UpMax: User partitioning for MaxSAT}

\author{Pedro Orvalho}{INESC-ID, Instituto Superior Técnico, Universidade de Lisboa, Portugal}{pmorvalho@tecnico.ulisboa.pt}{https://orcid.org/0000-0002-7407-5967}{}
\author{Vasco Manquinho}{INESC-ID, Instituto Superior Técnico, Universidade de Lisboa, Portugal}{vasco.manquinho@tecnico.ulisboa.pt}{https://orcid.org/0000-0002-4205-2189}{}
\author{Ruben Martins}{Carnegie Mellon University, USA}{rubenm@andrew.cmu.edu}{https://orcid.org/0000-0003-1525-1382}{}

\authorrunning{P. Orvalho, V. Manquinho, R. Martins} 

\Copyright{Pedro Orvalho and Vasco Manquinho and Ruben Martins} 



\category{} 

\relatedversion{} 

\funding{This work was partially supported by NSF award CCF-1762363 and an Amazon Research Award. This work was also supported by Portuguese national funds through FCT under projects UIDB/50021/2020, PTDC/CCI-COM/2156/2021, 2022.03537.PTDC and grant SFRH/\-BD/\-07724/\-2020, and by project ANI 045917 funded by FEDER and FCT.
}

\usepackage{graphicx}
\usepackage{tikz}
\usepackage{hyperref}
\usepackage{amsfonts}
\usepackage{comment}
\usepackage[linesnumbered,ruled,vlined]{algorithm2e}
\usepackage{amsmath}
\usepackage{mathtools}
\usepackage{rotating}
\usepackage{multirow}
\usepackage{booktabs}

\newcommand\bigforall{\mbox{\Large $\mathsurround0pt\forall$}} 
\newcommand{\openwbo}{\textsc{Open-WBO}\xspace}

\newcommand{\upmax}{\textsc{UpMax}\xspace}

\begin{document}
\maketitle              

\begin{abstract}
It has been shown that Maximum Satisfiability (MaxSAT) problem instances can 
be effectively solved by partitioning the set of soft clauses into several 
disjoint sets.
The partitioning methods can be based on clause weights
(e.g., stratification) or based on graph representations of
the formula. Afterwards, a merge procedure is applied to guarantee
that an optimal solution is found. 

This paper proposes a new framework called \upmax that decouples the partitioning
procedure from the MaxSAT solving algorithms. As a result, new partitioning
procedures can be defined independently of the MaxSAT algorithm to be
used. Moreover, this decoupling also allows users that build new MaxSAT
formulas to propose partition schemes based on knowledge of the problem 
to be solved. 
We illustrate this approach using several problems and show that 
partitioning has a large impact on the performance of unsatisfiability-based MaxSAT algorithms.
\end{abstract}

\section{Introduction}
\label{sec:intro}

In the last decade, Maximum Satisfiability (MaxSAT) algorithmic improvements
have resulted in the successful usage of MaxSAT algorithms in several
application domains such as fault localization~\cite{majumdar-pldi11},
scheduling~\cite{DBLP:journals/anor/DemirovicMW19}, 
planning~\cite{DBLP:conf/aaai/ZhangB12}, 
data analysis~\cite{DBLP:conf/sat/BergHJ18},
among other~\cite{PackUp12,DBLP:conf/confws/WalterZK13,DBLP:conf/dft/HosokawaYMYHA19}.
These improvements resulted from new algorithm designs~\cite{morgado-constraints13} based on iterative calls to a highly efficient Satisfiability (SAT) solver. 
However, MaxSAT algorithms also take advantage of 
other techniques, such as effective encodings of cardinality 
constraints~\cite{DBLP:journals/constraints/ZhaKF19} or the incremental 
usage of SAT solvers~\cite{martins-cp14}.

Another technique for MaxSAT solving is to use partitioning on the
soft clauses. For instance, several solvers use partitioning of soft
clauses according to their weight~\cite{ansotegui-cp12}, 
which are particularly effective when the MaxSAT instance encodes a 
lexicographic optimization problem~\cite{DBLP:journals/amai/Marques-SilvaAGL11}.
For the particular case of partial MaxSAT, other techniques have been proposed, such as using a graph representation of the formula~\cite{DBLP:conf/sat/NevesMJLM15}. However, despite its success for some classes of benchmarks, graph-based partitioning has not been widely used mainly because (1) the graph representation may become too large to build or to process, and (2) it is not decoupled with the base MaxSAT algorithm (i.e., changing the partition method implies altering the MaxSAT algorithm).
Furthermore, in some cases, the partitions might not capture the problem structure that is helpful for MaxSAT solving. Since several MaxSAT algorithms rely on identifying unsatisfiable subformulas, each partition should be an approximation of an unsatisfiable subformula to be solved separately.

\begin{figure}[t!]
    \centering
    \scalebox{0.35}{\includegraphics{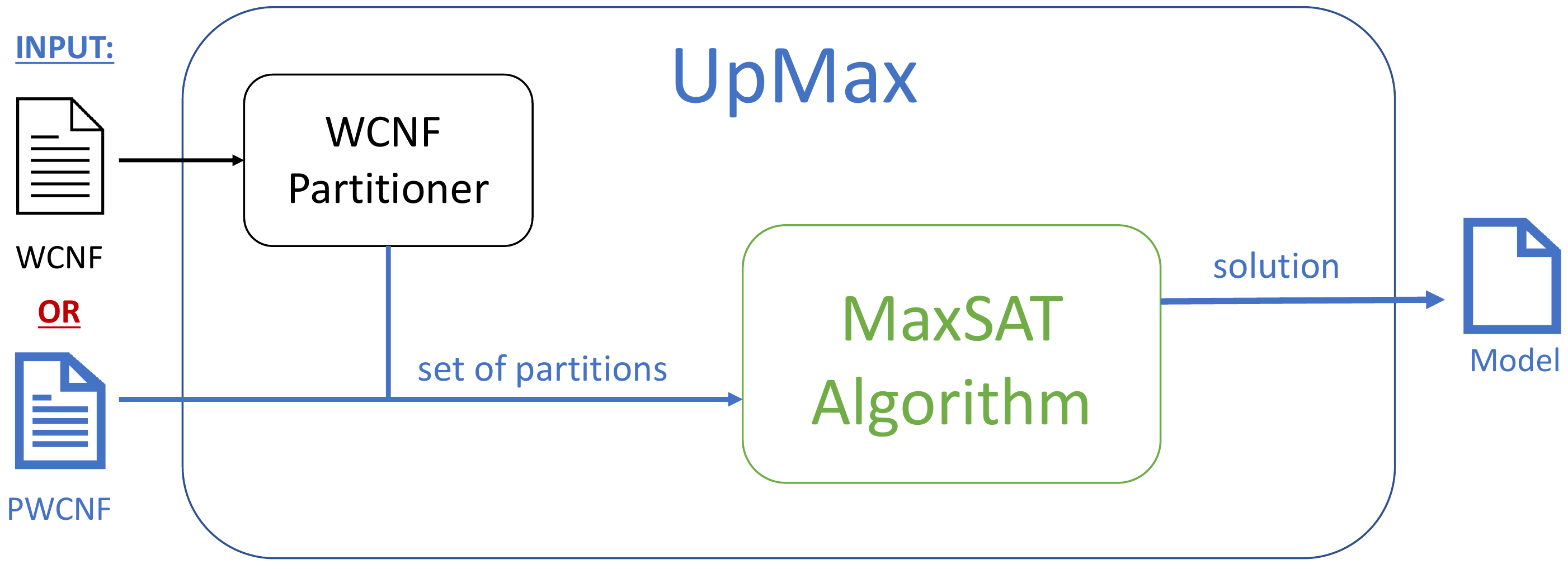}}
    \caption{Overview of UpMax.}
    \label{fig:UpMax-overview}
\end{figure}

Until now, the partitioning of MaxSAT formulas is interconnected to the subsequent algorithm to be used. Therefore, it is not easy to define and test new partitioning methods with several MaxSAT algorithms developed by different people. The first contribution of this work is to propose a new format called {\tt pwcnf} for defining MaxSAT formulas where clauses are split into partitions. 
Figure~\ref{fig:UpMax-overview} illustrates the schematic view of the \upmax architecture based on the decoupling of the MaxSAT solving algorithm from the split of the clauses on the MaxSAT formula. Observe that any partitioning method (e.g., graph-based partitioning~\cite{DBLP:conf/sat/NevesMJLM15}) can be used to generate the instances in the new {\tt pwcnf} format. Hence, this new format allows decoupling of the partitioning procedure from the MaxSAT algorithm, facilitating the appearance of new partition methods for MaxSAT formulas.
Secondly, \upmax is not restricted to any partitioning scheme. \upmax also allows the MaxSAT user to propose how to partition MaxSAT formulas based on her domain knowledge of the problem to be solved. Note that this is not possible with current MaxSAT tools.
Thirdly, with little effort, MaxSAT algorithms based on unsatisfiability approaches can be adapted to the new format. This is possible due to the newly proposed \upmax architecture. Hence, this paper presents the results of several algorithms using different partitioning schemes.
Finally, we present several use cases where different user-defined partitioning schemes can be easily defined and tested. Experimental results show that user-based partitioning significantly impacts the performance of MaxSAT algorithms. 
Thus, \upmax 
decouples clause partitioning from MaxSAT solving, opening new research directions for partitioning, modeling, and algorithm development.

\section{Background}
\label{sec:background}

A propositional formula in Conjunctive Normal Form (CNF) is defined as a conjunction of clauses where a clause is a disjunction of literals
such that a literal is either a propositional variable $v_i$ or its
negation $\neg v_i$.
Given a CNF formula $\phi$, the Satisfiability (SAT) problem corresponds 
to decide if there is an assignment such that $\phi$ is satisfied or
prove that no such assignment exists.
The Maximum Satisfiability (MaxSAT) is an optimization version of the
SAT problem. Given a CNF formula $\phi$, the goal is to find an assignment 
that minimizes the number of unsatisfied clauses in $\phi$. 
In partial MaxSAT, clauses in $\phi$ are split in hard $\phi_h$ and 
soft $\phi_s$. Given a formula $\phi = (\phi_h, \phi_s)$, the goal
is to find an assignment that satisfies all hard clauses in $\phi_h$
while minimizing the number of unsatisfied soft clauses in $\phi_s$. The partial MaxSAT problem can be further generalized to the weighted version, where each soft clause has an associated weight, and the optimization goal is to minimize the sum of the weights of the unsatisfied soft clauses.
Finally, we assume that $\phi_h$ is satisfiable. Moreover,
the set notation is also commonly used to manipulate formulas and clauses,
i.e., a CNF formula can be seen as a set of clauses (its conjunction), and a clause as a set of literals (its disjunction). 

\subsection{Algorithms for MaxSAT}
\label{sec:alg-maxsat}

Currently, state of the art algorithms for MaxSAT are based on successive 
calls to a SAT solver~\cite{morgado-constraints13}.
One of the approaches is to perform a SAT-UNSAT linear search on the number
of unsatisfied soft clauses. For that, the algorithm starts by adding a 
new relaxation variable $r_i$ to each soft clause $s_i \in \phi_s$,
where $r_i$ represents the unsatisfiability of clause $s_i$. 
Next, it defines an initial upper bound $\mu$ on the number of unsatisfied 
soft clauses. At each SAT call, the constraint $\sum r_i \le \mu -1$ is
added such that an assignment that improves on the previous one is
found.
Whenever the working formula becomes unsatisfiable, then the previous
SAT call identified an optimal assignment. There are a plethora of
these algorithms~\cite{qmaxsat-jsat12,DBLP:journals/constraints/ZhaKF19,DBLP:conf/sat/PaxianR018,piotrow2019uwrmaxsat}.

On the other hand, UNSAT-SAT algorithms start with a lower bound $\lambda$
on the number of unsatisfied soft clauses initialized at 0.
The algorithm starts with an overconstrained working formula. 
At each iteration, the working formula is relaxed by adding additional
relaxation variables allowing more soft clauses to be unsatisfied.
Whenever the working formula becomes satisfiable, then an optimal solution
is found. There are also many successful MaxSAT solvers that use an 
UNSAT-SAT 
approach~\cite{FM06,ansotegui-cp12,ansotegui-cp13,manquinho-sat09,wmsu3-corr07,morgado-cp14,davies-sat13}.
Two key factors for the performance of UNSAT-SAT algorithms is the
usage of SAT solvers to identify unsatisfiable subformulas, and
the search process being incremental~\cite{een-jsat06,martins-cp14}.
Instead of dealing with the whole formula at once, some algorithms
try to split the formula into partitions~\cite{ansotegui-cp12,martins-ecai12,DBLP:conf/sat/NevesMJLM15}. 
In particular, partitioning focuses on splitting the set of soft 
clauses into disjoint sets.
The motivation is to quickly identify a minimal cost considering 
just a subset of soft clauses. Since the sets are disjoint, the 
sum of the minimal cost of all partitions defines a lower bound 
on the optimal solution.
Moreover, a smaller instance should be able to be easier to solve.
Hence, the convergence to the optimum is expected to be faster.

\DontPrintSemicolon
\SetKwFunction{soft}{soft}
\SetKwFunction{SAT}{SAT}
\SetKwFunction{decomposeSoft}{partitionSoft}
\SetKwFunction{msu}{MaxSAT}
\SetKwFunction{selectPart}{selectPartitions}
\SetKwFunction{mergePart}{mergePartitions}
\SetKwFunction{first}{first}
\SetKwFunction{weight}{weight}
\SetKwFunction{encodeCNF}{CNF}
\SetKwFunction{min}{min}
\SetKwData{result}{satisfiable assignment to}
\SetKwData{unsat}{UNSAT}
\SetKwData{sat}{SAT}
\SetKwData{minc}{min$_\textnormal{c}$}
\SetKwData{true}{true}
\SetKwData{st}{st}
\SetVlineSkip{1pt}
\begin{algorithm}[!t]
  \small
  \KwIn{$\phi = (\phi_h, \phi_s)$}
  \KwOut{optimal assignment to $\phi$}
  $\gamma \gets \langle \gamma_1, \ldots, \gamma_n\rangle \gets \decomposeSoft(\phi_h, \phi_s)$\label{li:split}\tcp*[r]{\footnotesize initial partitions}
  \If{$|\gamma| = 1$}{
    \Return{\msu($\phi_h, \phi_s$)}\label{li:onePart}\tcp*[r]{\footnotesize no partitions}
  }
  \ForEach{$\gamma_i \in \gamma$}{\label{li:round1-begin}
    $\nu \gets \msu(\phi_h, \gamma_i)$\label{li:round1-end}\;
  }
  \While{\true}{
    $(\gamma_i, \gamma_j) \gets \selectPart(\gamma)$\label{li:selectPart}\;
    $\gamma_k \gets \mergePart(\gamma_i, \gamma_j)$\label{li:mergePart}\;
    $\gamma \gets \gamma \setminus \{ \gamma_i, \gamma_j \} \cup \{ \gamma_k \}$ \label{li:updateSet}\tcp*[r]{\footnotesize update partition set}
    $\nu \gets \msu(\phi_h, \gamma_k)$\label{li:msu3-K-begin}\;
    \If{$|\gamma| = 1$}{
      \Return{$\nu$}\label{li:return}
    }
  }
  \caption{Generic Partition-based MaxSAT Algorithm}\label{alg:part}
\end{algorithm}

Algorithm~\ref{alg:part} presents the pseudo-code for this generic 
approach. First, soft clauses in $\phi_s$ are split into $n$ disjoint
sets (line~\ref{li:split}). If $n = 1$, then there is no partitioning
and a MaxSAT solver is called on the whole formula. Otherwise, a
MaxSAT solver solves each partition $\gamma_i$ independently (line~\ref{li:round1-end}). Next, two partitions are selected and
merged~(lines~\ref{li:selectPart}-\ref{li:mergePart}). The newly
merged partition $\gamma_k$ is then solved considering
the information already obtained from solving $\gamma_i$ and $\gamma_j$.
This process is repeated until there is only one partition whose
solution is an optimal assignment to the original MaxSAT instance (line~\ref{li:return}).
Observe that several MaxSAT algorithms can
be used in this scheme including Fu-Malik~\cite{FM06}, WPM3~\cite{DBLP:journals/ai/AnsoteguiG17},
MSU3~\cite{wmsu3-corr07}, OLL~\cite{morgado-cp14} or a hitting set approach~\cite{davies-sat13,DBLP:conf/sat/SaikkoBJ16}, 
among others~\cite{morgado-constraints13}.

\begin{figure}[t]
  \centering
    \begin{tabular}{llllllll}
    Hard: & $h_1 : (v_1 \vee v_2)$           & & $h_2 : (\neg v_2 \vee v_3)$ &
          & $h_3 : (\neg v_1 \vee \neg v_3)$ & & $h_4 : (v_4 \vee v_5)$ \\
          & $h_5 : (\neg v_5 \vee v_6)$      & & $h_6 : (\neg v_4 \vee \neg v_6)$ &
          & $h_7 : (\neg v_3 \vee \neg v_6)$ \\
    Soft: & $s_1: (\neg v_1)$ & & $s_2: (\neg v_3)$ & 
          & $s_3: (\neg v_4)$ & & $s_4: (\neg v_6)$ 
    \end{tabular}
    \caption{Example of a MaxSAT formula}
    \label{fig:max-sat-formula}
\end{figure}

\begin{example}
Consider the MaxSAT formula in Figure~\ref{fig:max-sat-formula}.
Suppose the soft clauses are split into two disjoint sets $\gamma_1 = \{ s_1, s_2 \}$ and $\gamma_2 = \{ s_3, s_4 \}$. Next, a MaxSAT solver is applied to
MaxSAT instances $(\phi_h, \gamma_1)$ and $(\phi_h, \gamma_2)$. Each of 
these instances has an optimal solution of 1. When merging both partitions, 
a final MaxSAT call is made on $(\phi_h, \gamma_1 \cup \gamma_2)$ with an 
initial lower bound of 2 (because $\gamma_1$ and $\gamma_2$ are disjoint). 
Since the lower bound is already equal to the optimum value, this last 
call is not likely to be computationally hard. 
\end{example}

A related approach to partitioning is \textit{Group MaxSAT}~\cite{DBLP:journals/heuristics/ArgelichM06,herasMM12-groupMaxSAT}. Group MaxSAT is a variation of MaxSAT where soft clauses are grouped, and each group has a weight. The optimization goal in Group MaxSAT is to minimize the sum of the weights of the unsatisfied groups. A group is considered unsatisfied if at least one of its soft clauses is unsatisfied. Note that Group MaxSAT and MaxSAT are solving different optimization problems. The partitions in Algorithm~\ref{alg:part} do not change the optimization goal of MaxSAT but instead are meant to guide the solver to find an optimal solution to the MaxSAT formula.

\subsection{Partitioning MaxSAT Formulas}
\label{sec:partitions}

There are several graph representations for CNF formulas that
have been proposed in order to analyze its structural 
properties~\cite{yates-AI70,vangelder-cp11,ansotegui-sat12}. 
For instance, it is well-known that industrial SAT instances
can be represented in graphs with high modularity~\cite{ansotegui-sat12}.
On the other hand, graphs that represent randomly generated
instances are closer to an Erd\"os-R\'enyi model. Similar observations
have also been made in MaxSAT instances~\cite{DBLP:conf/sat/NevesMJLM15}.
Furthermore, based on these graph representations, one can partition
the set of soft clauses in a MaxSAT instance by applying a community
finding algorithm~\cite{blondel08} that maximizes the modularity value.

\begin{figure}[t]
  \centering
    \scalebox{0.7}{\begin{tikzpicture}[thick,xscale=.30]
  \tikzstyle{every node}=[circle,draw]
  \foreach \a in {1,3,6} \node (v\a) at (\a*3-3,2)  {$v_{\a}$} ;
  \foreach \a in {2,5,4} \node (v\a) at (\a*4-4,0)  {$v_{\a}$} ;
  \foreach \i in {2,3} \draw(v1)--(v\i);
  \foreach \i in {3} \draw(v2)--(v\i);
  \foreach \i in {6} \draw(v3)--(v\i);
  \foreach \i in {5,6} \draw(v4)--(v\i);
  \foreach \i in {6} \draw(v5)--(v\i);
  \foreach \a in {1,3} \node[fill=green] (v\a) at (\a*3-3,2)  {$v_{\a}$} ;
  \foreach \a in {2} \node[fill=green] (v\a) at (\a*4-4,0)  {$v_{\a}$} ;
  \foreach \a in {4,5} \node[fill=orange!75] (v\a) at (\a*4-4,0)  {$v_{\a}$} ;
  \foreach \a in {6} \node[fill=orange!75] (v\a) at (\a*3-3,2)  {$v_{\a}$} ;
\end{tikzpicture}} \hspace{1cm}
    \scalebox{0.7}{\begin{tikzpicture}[thick,xscale=1.5]
  \tikzstyle{every node}=[circle,draw]
  \node[fill=green](h1) at (0,1) {$h_1$};
  \node[fill=yellow!50](h2) at (1,1) {$h_2$};
  \node[fill=green](h3) at (0.5,2) {$h_3$};
  \node[fill=yellow!50](h7) at (2,1) {$h_7$};
  \node[fill=orange!75](h5) at (3,1) {$h_5$};
  \node[fill=orange!75](h4) at (4,1) {$h_4$};
  \node[fill=orange!75](h6) at (3.5,2) {$h_6$};
  \node[fill=green](s1) at (0,0) {$s_1$};
  \node[fill=yellow!50](s2) at (1,0) {$s_2$};
  \node[fill=orange!75](s4) at (3,0) {$s_4$};
  \node[fill=orange!75](s3) at (4,0) {$s_3$};
  \draw(h1)--(h2)--(h7)--(h5)--(h4);
  \draw(h1)--(h3);
  \draw(h2)--(h3);
  \draw(h5)--(h6);
  \draw(h4)--(h6);
  \draw(h1)--(s1);
  \draw(h2)--(s2);
  \draw(h5)--(s4);
  \draw(h4)--(s3);
\end{tikzpicture}}
  \caption{VIG graph (left) and RES graph (right) for MaxSAT formula in Figure~\ref{fig:max-sat-formula}}
  \label{fig:vig}
\end{figure}
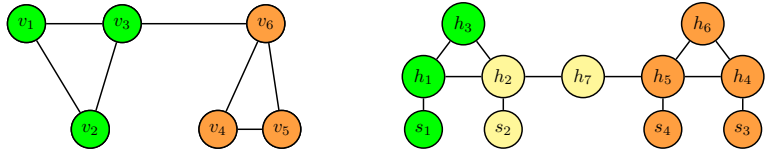

One possible graph representation is the Variable Incidence Graph (VIG).
Let $G = (V,E)$ denote a weighted undirected graph where $V$ defines
the graph vertices and $E$ its edges.
In the VIG representation, we have a vertex $v_i \in V$ for each 
variable $v_i$ in the MaxSAT formula $\phi$. Next, 
for each pair of variables $v_i$ and $v_j$, if there is at least one
clause in $\phi$ that contains both variables $v_i$ and $v_j$ 
(or its negated literals), then an edge $(v_i, v_j)$ is added to the graph.
For each clause $c \in \phi$ with $n$ literals, then $1/(^n _2)$ is
added to the weight of every pair of variables that occurs in clause $c$.
Figure~\ref{fig:vig} illustrates the VIG representation 
for the MaxSAT formula in Figure~\ref{fig:max-sat-formula}. Note that edge weights are 
not represented to simplify the figure. Next, if we apply a community
finding algorithm that maximizes the modularity, two
communities are identified (vertices with different colors).
Hence, soft clauses with variables in $\{v_1, v_2, v_3 \}$ would define
a partition, while soft clauses with variables in $\{v_4, v_5, v_6 \}$
define the other partition. Therefore, we would have $\gamma_1 = \{s_1, s_2 \}$
and $\gamma_2 = \{s_3, s_4 \}$ as the two partitions of soft clauses. 
Observe that $\phi_h \cup \gamma_1$ is an unsatisfiable subformula 
of $\phi$, as well as $\phi_h \cup \gamma_2$.
In the Clause-Variable Incidence Graph (CVIG) representation, there is a node for each variable 
and another node for each clause. Moreover, if a variable $v_i$ 
(or its negation $\neg v_i$) occurs in a clause $c_j$, then there is 
an edge $(v_i, c_j)$ in the graph. 

On the other hand, in Resolution-based Graphs (RES) only clauses are 
represented as vertices. Hence, for each clause $c_j$ there is a node 
in the graph.
Let $c^{r}_{jk}$ be the clause that results from applying the resolution
operation between clauses $c_j$ and $c_k$. If $c^{r}_{jk}$ is not a
tautology, then an edge $(c_j, c_k)$ is added to the graph with
weight $1/|c^{r}_{jk}|$ where $|c^{r}_{jk}|$ denotes the size of
the resolvent clause. Note that if the resolution operation
results in a trivial resolvent, then no edge is added.
The right graph in Figure~\ref{fig:vig} shows the RES graph 
representation of the MaxSAT formula in Figure \ref{fig:max-sat-formula}. 
Colors illustrate the three communities of soft clauses found in this formula, i.e., $\gamma_1 = \{ s_1 \}$, $\gamma_2 = \{ s_2 \}$ and
$\gamma_3 = \{ s_3, s_4 \}$.

\section{User-Based Partitioning}
\label{sec:use-case}

All the partitioning methods described in Section~\ref{sec:partitions} are automatic and attempt to recover some partition scheme from the structure of the MaxSAT formula. However, when encoding a problem into MaxSAT, it is often the case that the user has enough domain knowledge to provide a potential partition scheme. Unfortunately, the current MaxSAT format does not support this extra information. 
\label{sec:pwcnf}
Thus,
 we propose a new generic format for MaxSAT, \texttt{pwcnf}, where the partition scheme can be defined, and solvers can take advantage of it. 
The \texttt{pwcnf} format starts with a header:
\vspace*{-1.5mm}
\begin{verbatim}
p pwcnf n_vars n_clauses topw n_part
\end{verbatim}
\vspace*{-1.5mm}
and each line in the body is of the form: \texttt{[part] [weight] [literals*] 0}.

In the header, \texttt{n\_vars} and \texttt{n\_clauses} are the numbers of variables and clauses of the formula. \texttt{topw} is the weight assigned to the hard clauses and \texttt{n\_part} is the number of partitions. Each partition label (\texttt{[part]}) must be a positive integer from 1 to \texttt{n\_part}.

\begin{example}
Considering the RES graph presented in Figure \ref{fig:vig}, clearly it has 3 partitions accordingly to the coloring scheme: green (label 1), yellow (label 2) and orange (label 3). The \texttt{pwcnf} for this graph is the following:

\vspace{2mm}
\begin{tabular}{llll}
{\tt p pwcnf 6 11 7 3} \hspace{1cm} & {\tt 1 7 -1 -3 0 } & {\tt 3 7 -4 -6 0 } & {\tt 2 1 -3 0 }\\ 
{\tt 1 7 1 2 0 } & {\tt 3 7 4 5 0 } & {\tt 2 7 -3 -6 0 } & {\tt 3 1 -4 0 }\\
{\tt 2 7 -2 3 0 }  & {\tt 3 7 -5 6 0 } & {\tt 1 1 -1 0 } & {\tt 3 1 -6 0 }\\
\end{tabular}
\end{example}

Alloy~\cite{alloy} is a declarative modeling language based on a first-order relational logic that has been applied to different software engineering problems~\cite{khurshid-04,donoso-abz14,kang-fse16,Trippel2018}.
Recently, $\text{Alloy}^{\text Max}$ has been proposed that extends Alloy with the ability to find optimal solutions~\cite{fse21-AlloyMax}.
$\text{Alloy}^{\text Max}$ uses a MaxSAT solver as its optimization engine. So for that, $\text{Alloy}^{\text Max}$ encodes in a MaxSAT formula the high-level Alloy model.
Moreover, $\text{Alloy}^{\text Max}$ uses domain knowledge to partition the soft clauses in the generated MaxSAT formula. In particular, one partition of soft clauses is created for each optimization operator used in the Alloy specification.
$\text{Alloy}^{\text Max}$ was the first successful usage of \upmax, which shows that the creation of the new file format makes it easier for other researchers to integrate partitioning in their applications.

\subsection{Use Case: Minimum Sum Coloring}
\label{sec:mscp}

To illustrate how a user can take advantage of the new proposed format, we analyze the Minimum Sum Coloring (MSC) problem from graph theory.
MSC is the problem of finding a proper coloring while minimizing the sum of the colors assigned to the vertices.
In this problem, the following conditions must be met: (1) each vertex should be assigned a color; (2) each vertex is assigned at most one color; and (3) two adjacent vertices cannot be assigned the same color. The optimization goal is to minimize the number of different colors in the graph.
Let $V$ denote the set of vertices in the graph. Let $C$ denote the set of possible colors. Let $X^v_c$ be the Boolean that is assigned to 1 if color $c$ is assigned to vertex $v$. The goal is to maximize $\neg X^v_c$, and each soft clause is assigned the weight of $c$.
The user can potentially group these soft clauses
in two distinct ways: (1) variables that share the same color are grouped, or
(2) variables that share the vertex number are grouped.
See Appendix~\ref{sec:msc:encoding} for more details on the MaxSAT encoding.

\begin{example}
Assume a user wants to minimize the number of different colors needed to color a given graph $G$ such that two adjacent vertices cannot share the same color. $G$ has 4 vertices, $v_1, \ldots, v_4$, and the following set of edges $G_E=\{(v_1,v_2), (v_1,v_3), (v_2,v_3), (v_3,v_4)\}$. Furthermore, there are 4 different colors available $c_1, \ldots, c_4$.
When encoding the problem into \texttt{pwcnf} the user could provide either the following VERTEX-based or COLOR-based partition scheme:

\begin{center}
\resizebox{0.9\columnwidth}{!}{%
\begin{tikzpicture}
    \filldraw[fill=green!20, draw=green!60] (-0.5,-0.2) ellipse (0.9cm and 1.4cm);
    \filldraw[fill=yellow!20, draw=yellow!60] (1.5,-0.2) ellipse (0.9cm and 1.4cm);
    \filldraw[fill=orange!20, draw=orange!60] (3.5,-0.2) ellipse (0.9cm and 1.4cm);
    \filldraw[fill=red!20, draw=red!60] (5.5,-0.2) ellipse (0.9cm and 1.4cm);
    
    \filldraw[fill=green!20, draw=green!60] (8.5,-0.2) ellipse (0.9cm and 1.4cm);
    \filldraw[fill=yellow!20, draw=yellow!60] (10.5,-0.2) ellipse (0.9cm and 1.4cm);
    \filldraw[fill=orange!20, draw=orange!60] (12.5,-0.2) ellipse (0.9cm and 1.4cm);
    \filldraw[fill=red!20, draw=red!60] (14.5,-0.2) ellipse (0.9cm and 1.4cm);

    \node at (2.5,2.2) {\emph{VERTEX-based}};
    \node at (-0.5,1.5) {$V_1$};
    \node at (1.5,1.5) {$V_2$};
    \node at (3.5,1.5) {$V_3$};
    \node at (5.5,1.5) {$V_4$};
    
    \node (x1) at (7,2.5) {};
    \node (x2) at (7,-2.5) {};
    \draw[thick] (x1) -- (x2);
    
    \node at (11.5,2.2) {\emph{COLOR-based}};
    \node at (8.5,1.5) {$C_1$};
    \node at (10.5, 1.5) {$C_2$};
    \node at (12.5,1.5) {$C_3$};
    \node at (14.5,1.5) {$C_4$};

    \node at (-0.5,0.7) {$\neg X_{v_1}^{c_1}$};
    \node at (-0.5,0.1) {$\neg X_{v_1}^{c_2}$};
    \node at (-0.5,-0.5) {$\neg X_{v_1}^{c_3}$};
    \node at (-0.5,-1.1) {$\neg X_{v_1}^{c_4}$};
    \node at (1.5,0.7) {$\neg X_{v_2}^{c_1}$};
    \node at (1.5,0.1) {$\neg X_{v_2}^{c_2}$};
    \node at (1.5,-0.5) {$\neg X_{v_2}^{c_3}$};
    \node at (1.5,-1.1) {$\neg X_{v_2}^{c_4}$};
    \node at (3.5,0.7) {$\neg X_{v_3}^{c_1}$};
    \node at (3.5,0.1) {$\neg X_{v_3}^{c_2}$};
    \node at (3.5,-0.5) {$\neg X_{v_3}^{c_3}$};
    \node at (3.5,-1.1) {$\neg X_{v_3}^{c_4}$};
    \node at (5.5,0.7) {$\neg X_{v_4}^{c_1}$};
    \node at (5.5,0.1) {$\neg X_{v_4}^{c_2}$};
    \node at (5.5,-0.5) {$\neg X_{v_4}^{c_3}$};
    \node at (5.5,-1.1) {$\neg X_{v_4}^{c_4}$};
    
    \node at (8.5,0.7) {$\neg X_{v_1}^{c_1}$};
    \node at (8.5,0.1) {$\neg X_{v_2}^{c_1}$};
    \node at (8.5,-0.5) {$\neg X_{v_3}^{c_1}$};
    \node at (8.5,-1.1) {$\neg X_{v_4}^{c_1}$};
    \node at (10.5,0.7) {$\neg X_{v_1}^{c_2}$};
    \node at (10.5,0.1) {$\neg X_{v_2}^{c_2}$};
    \node at (10.5,-0.5) {$\neg X_{v_3}^{c_2}$};
    \node at (10.5,-1.1) {$\neg X_{v_4}^{c_2}$};
    \node at (12.5,0.7) {$\neg X_{v_1}^{c_3}$};
    \node at (12.5,0.1) {$\neg X_{v_2}^{c_3}$};
    \node at (12.5,-0.5) {$\neg X_{v_3}^{c_3}$};
    \node at (12.5,-1.1) {$\neg X_{v_4}^{c_3}$};
    \node at (14.5,0.7) {$\neg X_{v_1}^{c_4}$};
    \node at (14.5,0.1) {$\neg X_{v_2}^{c_4}$};
    \node at (14.5,-0.5) {$\neg X_{v_3}^{c_4}$};
    \node at (14.5,-1.1) {$\neg X_{v_4}^{c_4}$};

\end{tikzpicture}
}
\vspace*{-3mm}
\end{center}
\end{example}

\subsection{Use Case: Seating Assignment Problem}
\label{sec:seating}

Another example of how a user can take advantage of the \texttt{pwcnf} format is the seating assignment problem. We encode this problem into MaxSAT  and show different partition schemes that can be provided by the user. Consider a seating assignment problem where the goal is to seat persons at tables such that the following properties are met: (1) Each table has a minimum and a maximum number of persons; (2) Each person is seated at exactly one table; and (3) Each person has some tags that represent their interests.
The optimization goal is to minimize the number of different tags between all persons seated at the same table. 

More formally, consider a seating problem with $p$ persons and $t$ tables.  Assume that the set of tags each person may have is defined by $G$ and the set of tables is defined by $T$. Consider the Boolean variables $Y_t^g$ that are assigned to 1 if there is \emph{at least one person} $p$ with a tag $g$ that is seated at table $t$. 
The goal of this optimization problem is to minimize $\sum_{{t\in T},{g \in G}} Y_t^g$ subject to the constraints of the problem.
When encoding the problem into MaxSAT, the $\neg Y_t^g$ literals will correspond to unit soft clauses. The user can potentially group these soft clauses in two distinct ways: (1) variables that share the same tag are grouped, or (2) variables that share the same table are grouped. We call the former partition scheme \emph{TAGS-based} and the latter \emph{TABLES-based}. 
More details 
can be found in Appendix~\ref{sec:assignment:encoding}.

\begin{example}
Consider that a user wants to seat 5 persons, $p_1, \ldots, p_5$, in two tables $t_1, t_2$. Each table must have at least 2 persons and at most 3 persons. Each person has a set of interests described by their tags as follows: 
\[
p_1 = \{A, B\}, p_2 = \{C\},p_3 = \{B\}, p_4 = \{C,A\}, p_5 = \{A\}
\]

When encoding the problem into \texttt{pwcnf} the user could provide either the following TAGS-based or TABLES-based partition scheme:
\begin{center}
\vspace*{-3mm}
\resizebox{0.65\columnwidth}{!}{%
\begin{tikzpicture}
    \filldraw[fill=green!20, draw=green!60] (-1.5,0) ellipse (0.8cm and 1.2cm);
    \filldraw[fill=yellow!20, draw=yellow!60] (0.5,0) ellipse (0.8cm and 1.2cm);
    \filldraw[fill=orange!20, draw=orange!60] (2.5,0) ellipse (0.8cm and 1.2cm);
    
    \filldraw[fill=green!20, draw=green!60] (5.5,0) ellipse (0.8cm and 1.2cm);
    \filldraw[fill=yellow!20, draw=yellow!60] (7.5,0) ellipse (0.8cm and 1.2cm);
    
    \node at (0.5,2.2) {\emph{TAGS-based}};
    \node at (-1.5,1.5) {$A$};
    \node at (0.5,1.5) {$B$};
    \node at (2.5,1.5) {$C$};
    
    \node (x1) at (4,2.5) {};
    \node (x2) at (4,-1.5) {};
    \draw[thick] (x1) -- (x2);
    
    \node at (6.5,2.2) {\emph{TABLES-based}};
    \node at (5.5,1.5) {$t_1$};
    \node at (7.5,1.5) {$t_2$};

    \node at (-1.5,0.5) {$\neg Y_{t_1}^A$};
    \node at (-1.5,-0.3) {$\neg Y_{t_2}^A$};
    \node at (0.5,0.5) {$\neg Y_{t_1}^B$};
    \node at (0.5,-0.3) {$\neg Y_{t_2}^B$};
    \node at (2.5,0.5) {$\neg Y_{t_1}^C$};
    \node at (2.5,-0.3) {$\neg Y_{t_2}^C$};
    
    \node at (5.5,0.7) {$\neg Y_{t_1}^A$};
    \node at (5.5,0) {$\neg Y_{t_1}^B$};
    \node at (5.5,-0.7) {$\neg Y_{t_1}^C$};
    \node at (7.5,0.7) {$\neg Y_{t_2}^A$};
    \node at (7.5,0) {$\neg Y_{t_2}^B$};
    \node at (7.5,-0.7) {$\neg Y_{t_2}^C$};
\end{tikzpicture}
}
\vspace*{-3mm}
\end{center}
\end{example}

\section{Experimental Results}
\label{sec:results}

\upmax is built on top of the open-source \openwbo MaxSAT solver~\cite{martins-sat14}. \upmax supports the new format \texttt{pwcnf} for user-based partitioning presented in Section~\ref{sec:use-case}. Alternatively, it can also take as input a \texttt{wcnf} formula and output a \texttt{pwcnf} formula using an automatic partitioning strategy based on VIG~\cite{martins-sat13}, CVIG~\cite{martins-sat13}, RES~\cite{DBLP:conf/sat/NevesMJLM15}, or randomly splitting the formula into $k$ partitions. \upmax can also be extended to support additional partitioning strategies that users may want to implement to evaluate their impact on the performance of MaxSAT algorithms. Furthermore, we are currently merging the partitions based on their size. We sort the partitions in increasing order, and we start by giving the smaller partition to the solver. However, other merging methods can be easily implemented and evaluated.

\upmax currently supports three UNSAT-based algorithms (WBO~\cite{manquinho-sat09}, OLL~\cite{morgado-cp14}, and MSU3~\cite{martins-cp14}) for both unweighted and weighted problems that take advantage of the partitions using the basic algorithm described in Algorithm~\ref{alg:part}.
WBO uses only at-most-one cardinality constraints when relaxing the formula at each iteration. In contrast, MSU3 uses a single cardinality constraint, and OLL uses multiple cardinality constraints. 
Furthermore, we have also extended RC2~\cite{imms19-RC2} and Hitman~\cite{mck13-HSEnum}, available in PySAT~\cite{imms18-PySAT}, to take advantage of user-based or graph-based partitions through our \texttt{pwcnf} formulae using Algorithm~\ref{alg:part}. RC2 is an improved version of the OLL algorithm~\cite{morgado-cp14,mims14-MSCG}. Hitman is a SAT-based implementation of an implicit minimal hitting set enumerator~\cite{mck13-HSEnum} and can be used as the basic flow of a MaxHS-like algorithm~\cite{db11-MaxHS} for MaxSAT. \upmax is publicly available at GitHub~\cite{UpMax}.

To show the impact of partitioning on the performance of unsatisfiability-based algorithms, we randomly generate 1,000 instances for both the seating assignment and the minimum sum coloring problem by varying the different parameters of each problem. 
We considered the automatic partitioning strategies available in \upmax (VIG, CVIG, RES, random; for the random partitioning strategy, we fixed $k = 16$), the user partitions (UP) described in Sections~\ref{sec:mscp} and~\ref{sec:seating}, and no partitions.
All of the experiments were run on \textsf{StarExec}~\cite{starexec} with a timeout of 1800 seconds and a memory limit of 32 GB for each instance.

\begin{table}[t!]
\centering
\caption{Number of solved instances for the Minimum Sum Coloring (MSC) problem.}
\label{tab:msc}
\resizebox{0.7\columnwidth}{!}{%
\begin{tabular}{@{}cccccccc@{}}
\cline{3-7}
\multicolumn{1}{l}{} &
  \multicolumn{1}{l|}{} &
  \multicolumn{2}{c|}{\textbf{User Part.}} &
  \multicolumn{3}{c|}{\textbf{Graph Part.}} &
  \multicolumn{1}{c}{} \\ \hline
\multicolumn{1}{|c|}{\textbf{Solver}} &
  \multicolumn{1}{c|}{\textbf{No Part.}} &
  \multicolumn{1}{c|}{\textbf{Vertex}} &
  \multicolumn{1}{c|}{\textbf{Color}} &
  \multicolumn{1}{c|}{\textbf{VIG}} &
  \multicolumn{1}{c|}{\textbf{CVIG}} &
  \multicolumn{1}{c|}{\textbf{RES}} &
  \multicolumn{1}{c|}{\textbf{Random}} \\ \hline
\textbf{MSU3}   & 245        & 758           & 770                    & 774      & 770               & 775               & \textbf{776} \\ \hline
\textbf{OLL}    & 796                 & 863           & 594                    & 945      & 944               & \textbf{947}      & 756          \\ \hline
\textbf{WBO}    & 483                   & 622                    & 314           & 745      & 750               & \textbf{755}      & 493          \\ \hline
\textbf{Hitman} & 610                   & 613                    & 471           & 605     & \textbf{614}      & 609               & 592          \\ \hline
\textbf{RC2}    & 796                  & 866                    & 528           & 943      & 939               & \textbf{944}      & 687          \\ \hline
\end{tabular}%
}
\end{table}

\begin{table}[t!]
\centering
\caption{Number of solved instances for the Seating Assignment problem.}
\label{tab:seating}
\resizebox{0.7\columnwidth}{!}{%
\begin{tabular}{@{}cccccccc@{}}
\cline{3-7}
\multicolumn{1}{l}{} &
  \multicolumn{1}{l|}{} &
  \multicolumn{2}{c|}{\textbf{User Part.}} &
  \multicolumn{3}{c|}{\textbf{Graph Part.}} &
  \multicolumn{1}{c}{} \\ \hline
\multicolumn{1}{|c|}{\textbf{Solver}} &
  \multicolumn{1}{c|}{\textbf{No Part.}} &
  \multicolumn{1}{c|}{\textbf{Table}} &
  \multicolumn{1}{c|}{\textbf{Tag}} &
  \multicolumn{1}{c|}{\textbf{VIG}} &
  \multicolumn{1}{c|}{\textbf{CVIG}} &
  \multicolumn{1}{c|}{\textbf{RES}} &
  \multicolumn{1}{c|}{\textbf{Random}} \\ \hline
\textbf{MSU3}   & 558                  & \textbf{671}           & 639                    & 659            & 641            & 640            & 565 \\ \hline
\textbf{OLL}    & 526                   & \textbf{634}           & 624                    & 627            & 599            & 608            & 528 \\ \hline
\textbf{WBO}    & 306                   & 400                    & \textbf{536}           & 400            & 385            & 386            & 360 \\ \hline
\textbf{Hitman} & 420                   & 403                    & \textbf{510}           & 406            & 425            & 420            & 440 \\ \hline
\textbf{RC2}    & 530                   & 620                    & \textbf{624}           & 618            & 600            & 597            & 541 \\ \hline
\end{tabular}%
}
\end{table}

\subsection{Minimum Sum Coloring Problem}
\label{sec:results_msc}

Table~\ref{tab:msc} shows the number of instances solved for the Minimum Sum Coloring problem for each partitioning strategy and algorithm. Entries highlighted in bold correspond to the highest number of instances solved for each algorithm.
The diversity of algorithms and partitions used allows us to make some interesting observations regarding the impact of partitioning on the performance of MaxSAT algorithms.
First, we can see that partitioning MaxSAT algorithms can often significantly outperform their non-partitioning counterparts. For instance, with partitioning the WBO algorithm can solve 272 more instances than without partitioning.
Secondly, most partition schemes result in performance improvements, even if the partition is done randomly. 
In this problem, random partitions had a benefit for the MSU3 
algorithm. This occurs since, until the last partition is added, this algorithm deals with a subset of soft clauses, resulting in finding smaller unsatisfiable cores.
Another observation is that the user-based partitions were not as good as the graph-based partitions. This may be partially explained by the fact that the user-based partitions do not consider the weight of the soft clauses which is important for weighted MaxSAT algorithms.
Finally, we can also observe that different partition 
strategies have different performance impacts on different algorithms. This suggests that new algorithms could leverage the partition information better than our approach presented in Algorithm~\ref{alg:part}. 

\begin{figure*}[t!]
    \begin{subfigure}[t!]{0.43\textwidth}
         \resizebox{\columnwidth}{!}{\includegraphics[width=\textwidth]{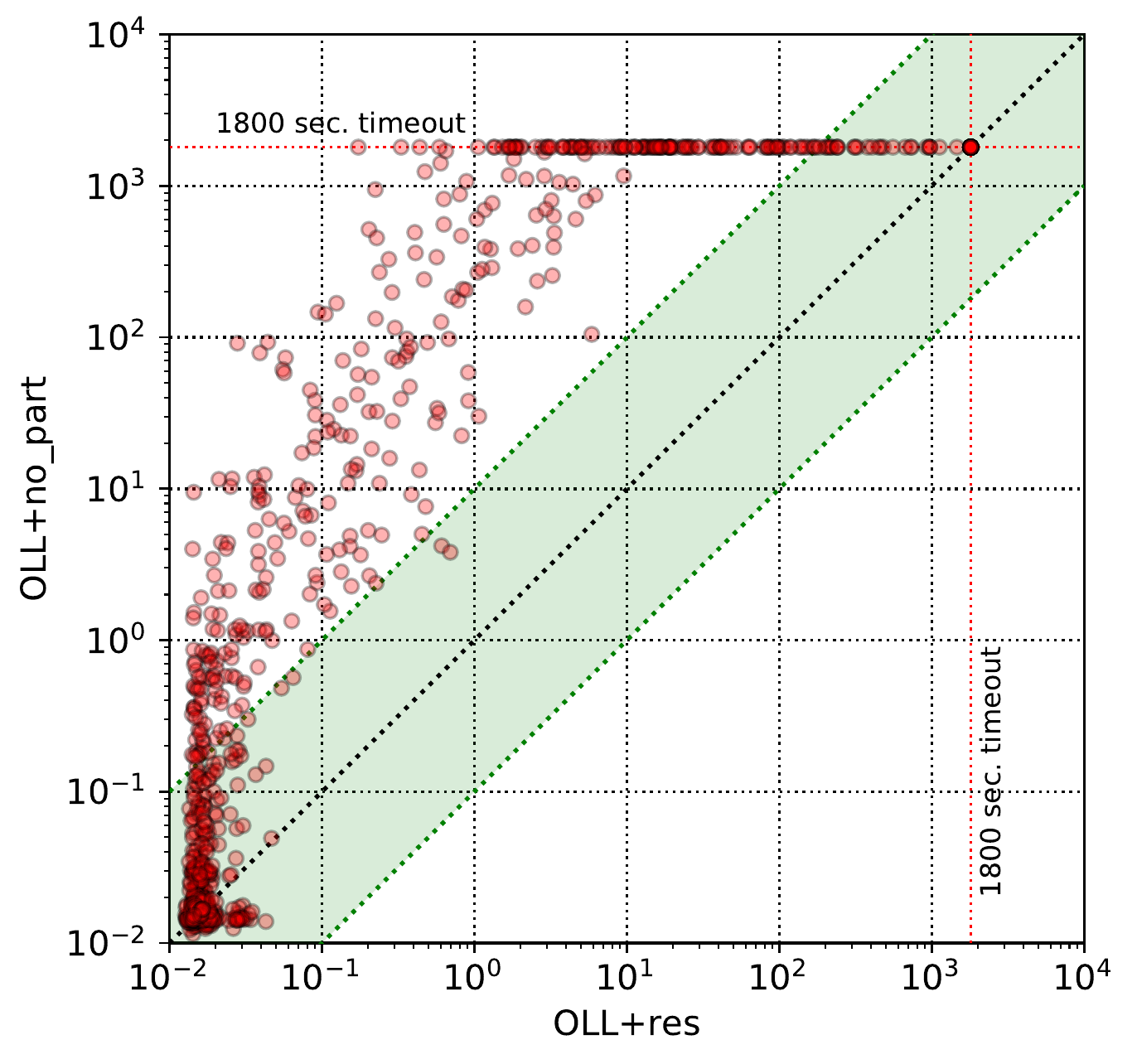}}
         \caption{MSC - OLL RES VS No Part.}
         \label{fig:msc-scatter-oll}
     \end{subfigure}
     \hfill
    \begin{subfigure}[t!]{0.43\textwidth}
    \centering
         \resizebox{\columnwidth}{!}{\includegraphics[width=\textwidth]{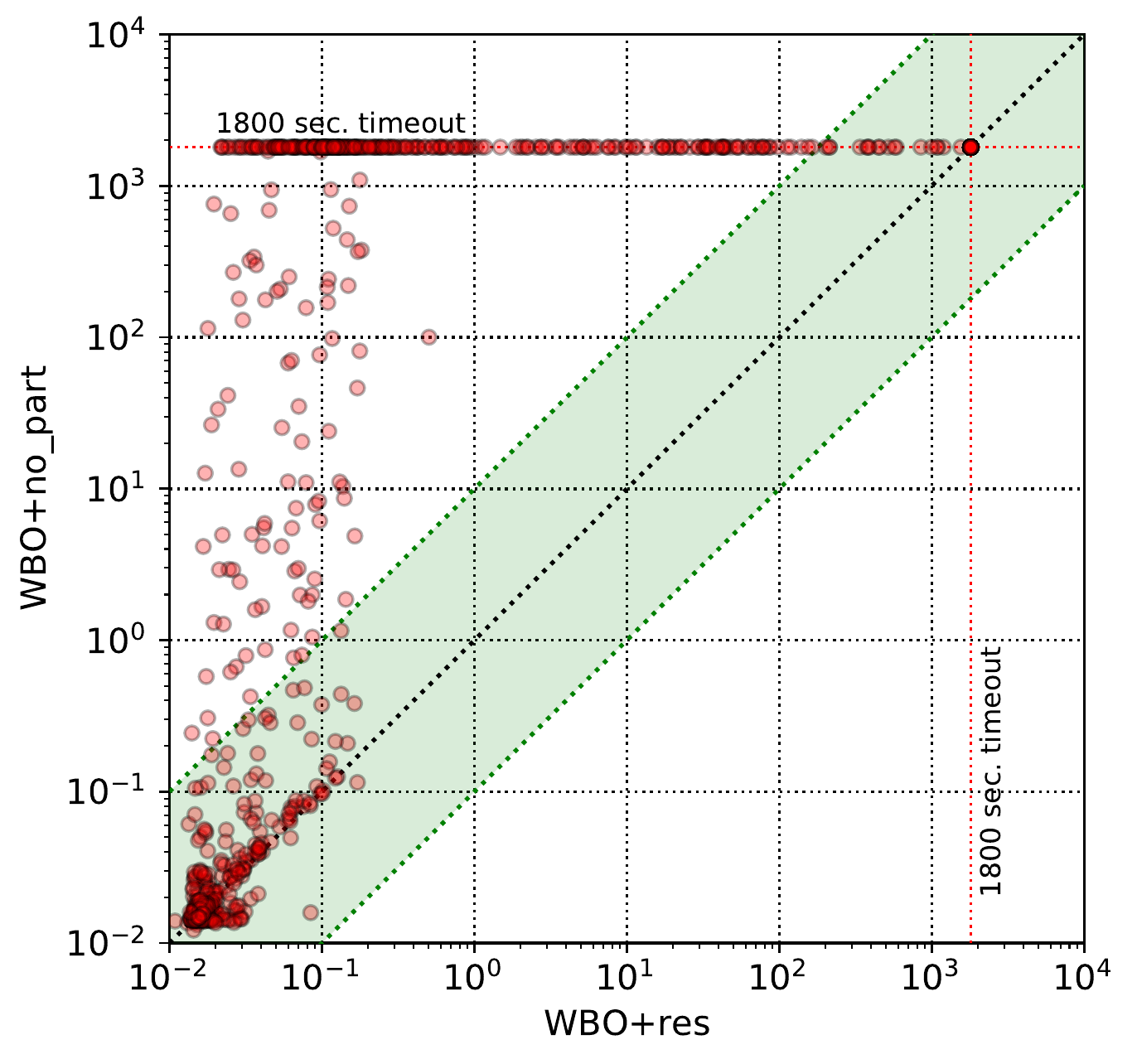}}
        \caption{MSC - WBO RES VS No Part.}
         \label{fig:msc-scatter-wbo}
     \end{subfigure}
    \begin{subfigure}[t!]{0.43\textwidth}
    \centering
         \resizebox{\columnwidth}{!}{\includegraphics[width=\textwidth]{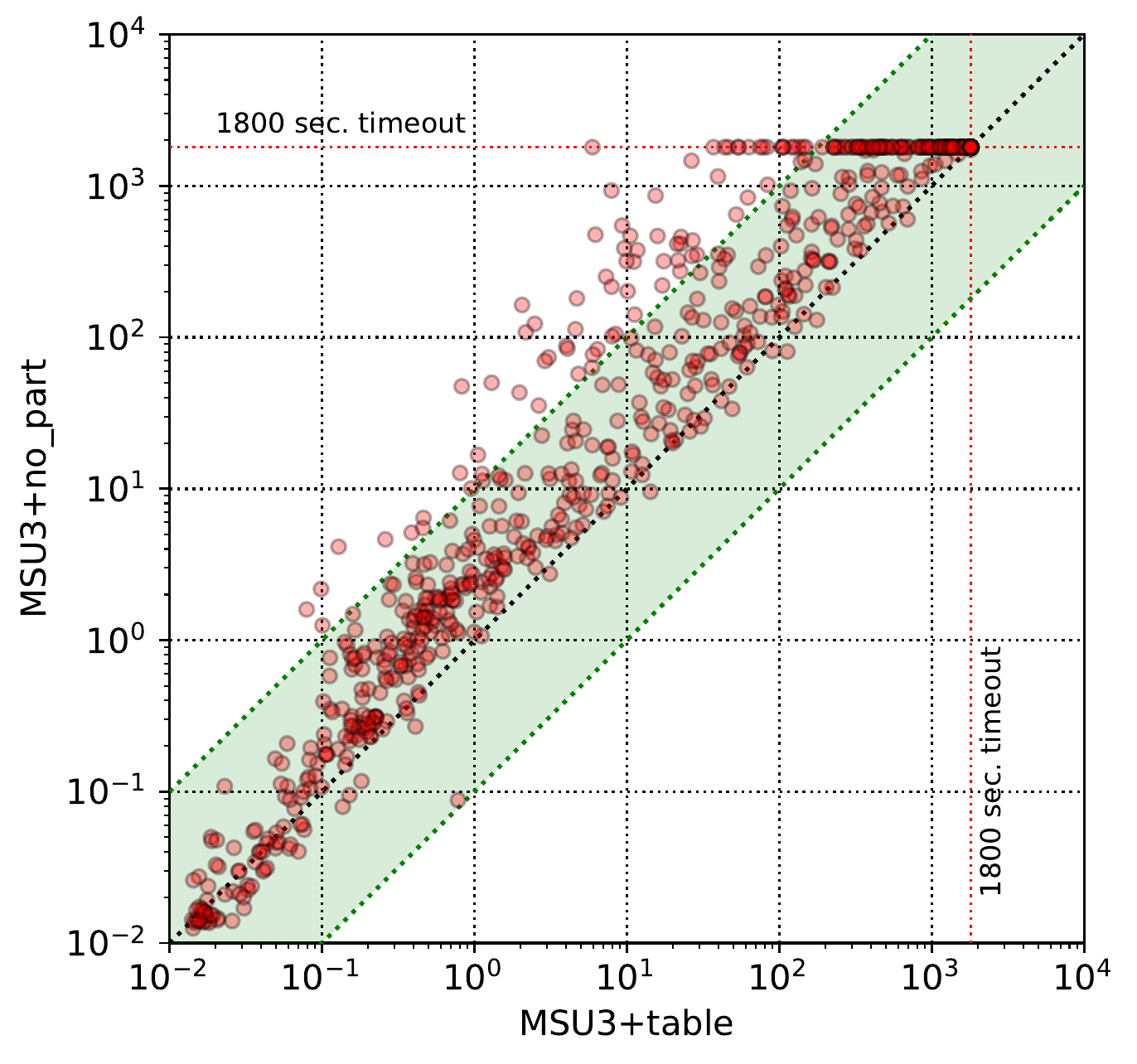}}
         \caption{SA - MSU3 Table VS No Part.}
         \label{fig:seating-scatter-msu3}
     \end{subfigure}
    \hfill
    \begin{subfigure}[t!]{0.43\textwidth}
    \centering
     \resizebox{\columnwidth}{!}{\includegraphics[width=\textwidth]{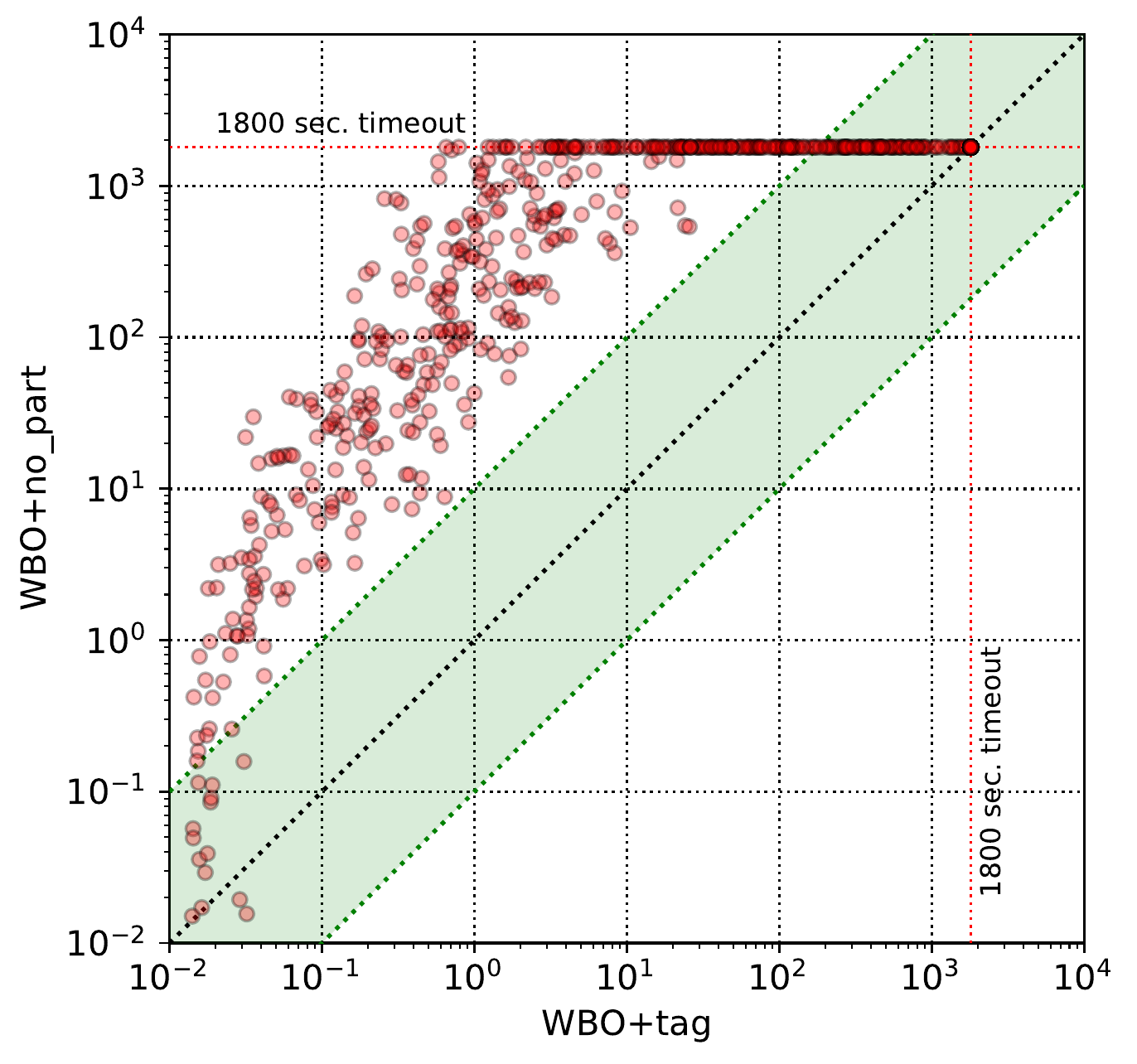}}
         \caption{SA - WBO Tag VS No Part.}
     \label{fig:seating-scatter-wbo}
    \end{subfigure} 
    \caption{Scatter plots comparing different MaxSAT algorithms and respective partitioning schemes for the problems of Minimum Sum Coloring (MSC) and Seating Assignment (SA).
    }
    \label{fig:scatter-plots}
\end{figure*}

Figures~\ref{fig:msc-scatter-oll}~and~\ref{fig:msc-scatter-wbo} show two scatter plots comparing two MaxSAT algorithms, OLL and WBO, on the set of instances for the Minimum Sum Coloring problem.
Each point in Figure~\ref{fig:msc-scatter-oll} represents an instance where the $x$-value (resp. $y$-value) is the CPU time spent to solve the instance using the OLL algorithm with the RES partitioning scheme (resp. OLL without any partitioning scheme). If a point is above the diagonal, then it means that the algorithm with partitioning outperformed the algorithm without partitioning. The OLL algorithm was the one with the best performance in the Minimum Sum Coloring (MSC) set of instances (see Table~\ref{tab:msc}). For many instances, we can observe a 10$\times$ speedup for OLL-RES when compared to OLL-NoPart. 
Secondly, Figure~\ref{fig:msc-scatter-wbo} compares the WBO algorithm with no partitioning and WBO using the RES partitioning scheme. WBO has the most significant gap between using partitions (755 instances solved) and not using partitions (483 instances solved). Figure~\ref{fig:msc-scatter-wbo} also shows that many instances that could not be solved without partitions can now be solved in a few seconds.
Both plots support that OLL and WBO greatly improve their performance on this set of benchmarks when using partitioning.
Appendix~\ref{appx:plots} presents cactus plots of our experiments.

\subsection{Seating Assignment Problem}
\label{sec:results-seating}

Table~\ref{tab:seating} shows the number of instances solved for the Seating Assignment problem for each partitioning strategy and algorithm. 
Similar to the previous use case, 
we can observe that partitioning can significantly impact the number of solved instances for all evaluated MaxSAT algorithms. For instance, WBO algorithm with partitioning can solve more 220 instances than without partitioning.
Furthermore, almost all the presented partition schemes, except random partitioning, result in performance improvements when compared to no partitioning scheme. 
Moreover, different partition strategies have different performance impacts on different algorithms. In this problem, user-based partitions achieved the best results. However, in some algorithms (e.g., MSU3, OLL), tabled-based partitioning is the best approach, while tag-based partitioning is better for the other algorithms.

Figures~\ref{fig:seating-scatter-msu3}~and~\ref{fig:seating-scatter-wbo} show two scatter plots comparing two MaxSAT algorithms, MSU3 and WBO, on the set of instances for the Seating Assignment problem.
Figure~\ref{fig:seating-scatter-msu3} shows the effectiveness of the table-based partitioning scheme against not using partitions in the MSU3 algorithm. We can observe that partitioning leads to faster runtimes, with most points being above the diagonal.
Secondly, Figure~\ref{fig:seating-scatter-wbo} compares the WBO algorithm with the tag-based and without any partitioning scheme. This algorithm has the biggest gap between using partitions (536 instances solved) and not using partitions (306 instances solved). 
Figure~\ref{fig:seating-scatter-wbo} supports that using partitioning with the WBO algorithm on this set of benchmarks greatly improves its performance with speedups of more than $10 \times$ for most of the instances.

\subsection{State-of-the-art MaxSAT Solvers}
\label{sec:other-solvers}

Using the \texttt{wcnf} formulae (No Part.) of both benchmark sets, we compared the performance of \upmax with some of the best solvers~\footnote{We did not modify any of these solvers since each solver has a large codebase with many optimizations, but these solvers could also use a partitioning approach.} in the MaxSAT Evaluation 2022~\cite{mse22}, such as MaxHS~\cite{davies-cp13}, UWrMaxSat-SCIP~\cite{DBLP:conf/ictai/Piotrow20}, CASHWMaxSAT-CorePlus~\cite{cashwmaxsat-21}, EvalMaxSAT~\cite{evalMaxSAT-20}, and MaxCDCL~\cite{maxCDCL-22}.
MaxHS is a MaxSAT solver based on an implicit hitting set approach. UWrMaxSat is an unsatisfiability-based solver using the OLL algorithm. These solvers can be seen as better versions than the RC2 and Hitman algorithms available in PySAT.
CASHWMaxSAT is developed   from UWrMaxSat, EvalMaxSAT is based on the OLL algorithm, and MaxCDCL is an extension for MaxSAT of
the CDCL algorithm~\cite{CDCL-4-MaxSAT}, which combines Branch and Bound and
clause learning.

Regarding the minimum sum coloring problem, MaxHS solved 873 instances, EvalMaxSAT solved 729, CASHWMaxSAT solved 993 (708 without SCIP), UWrMaxSat solved 994 (728 without SCIP), and MaxCDCL solved 995 instances. 
Note that solvers using Branch and Bound excel on these instances, and the performance of CASHWMaxSAT and UWrMaxSat deteriorates when SCIP is not used. Moreover, these results also show that partitioning improves less effective MaxSAT algorithms to become competitive with some solvers (e.g. UWrMaxSat),  and outperform other solvers, e.g., MaxHS and EvalMaxSAT.
Secondly, regarding the seating assignment problem, UWrMaxSat solved 580 instances, CASHWMaxSAT solved 585, MaxCDCL solved 593, MaxHS solved 643, and EvalMaxSAT solved 653 instances. When compared with our best results, note that table-based partitioning with the MSU3 algorithm can outperform all these solvers.
Moreover, since partitioning can improve the performance of multiple MaxSAT algorithms and it is beneficial for implicit hitting set approaches like Hitman, it has the potential to further improve the performance of MaxHS.

Even though partitioning is not expected to improve the performance of MaxSAT solvers on all problem domains, there are many domains similar to the seating assignment and minimum sum coloring~\footnote{Check $\text{Alloy}^\text{Max}$~\cite{fse21-AlloyMax} paper for \upmax's results on other application domains.} 
for which partitioning can provide a significant performance boost in MaxSAT solving. Finally, we note this work opens new lines of research based on decoupling of MaxSAT solvers from the procedure that defines the partitions of MaxSAT formulae.

\section{Conclusions}
\label{sec:conc}

In this paper, we propose \upmax, a new framework that decouples the partition generation from
the MaxSAT solving. 
\upmax allows the user to specify how to partition MaxSAT formulas 
with the proposed {\tt pwcnf} format. With this format, the partitioning of MaxSAT instances can be done a priori to MaxSAT solving.
Experimental results with two use cases with multiple algorithms show that partitioning can improve the performance of MaxSAT algorithms and allow them to solve more instances.
\upmax provides an extendable framework that can benefit (1) researchers on partitioning strategies, (2) solver developers with new MaxSAT algorithms that can leverage partition information, and (3) users that can benefit from additional information when modeling problems to MaxSAT.



\bibliography{mybibliography}

\newpage

\appendix

\section{Minimum Sum Coloring MaxSAT encoding}
\label{sec:msc:encoding}

In this section, we present the MaxSAT encoding used for the  Minimum Sum Coloring (MSC) problem, where the goal is to 
find a proper coloring while minimizing the sum of the colors assigned to the graph's vertices, such that the following properties hold:

\begin{itemize}
    \item Each vertex should be assigned a color;
    \item Each vertex is assigned at most one color;
    \item Two adjacent vertices cannot be assigned the same color;
    \item The optimization goal is to minimize the number of different colors in the graph.
\end{itemize}

\subsection*{Variables}  Let $V$ denote the set of vertices in the graph. Let $E$ denote the set of edges in the graph. Let $C$ denote the set of possible colors. Let $X^v_c$ be a Boolean variable that is assigned to True if color $c$ is assigned to vertex $v$. The goal is to maximise $\neg X^v_c$, and each soft clause is assigned the weight of $c$. The weight of each color $c$ is the index of this color on the set $C$.

\subsection*{Optimization criteria (weighted)} The goal of this optimization problem is to minimize $\sum_{v\in V} \sum_{c \in C}  c \cdot X_v^c$ subject to the constraints of the problem. Observe that usually, the cost associated with each color $c$ is the index of this color on the set $C$.

\subsection*{Constraints} We use the $X$ variables to encode the following hard constraints.

\begin{itemize}
    \item Each vertex $v$ should be assigned a color;
    \begin{itemize}
        \item ${\bigforall_{v \in V}} X_v^1 \vee X_v^2 \vee \dots \vee X_v^C$
    \end{itemize}
    \item Each vertex should be assigned at most one color;
    \begin{itemize}
        \item ${\bigforall_{v \in V}} \bigforall_{c\in C} \bigforall_{k\in {c+1 \dots C}} \neg X_v^c \vee \neg X_v^k$
    \end{itemize}
    \item Each two adjacent vertices $v$ and $u$ cannot be assigned the same color;
    \begin{itemize}
        \item ${\bigforall_{(v,u) \in E}}\bigforall_{c\in C} \neg X_v^c \vee \neg X_u^c$
    \end{itemize}
\end{itemize}

\begin{example}
Assume the Bob wants to minimize the number of different colors needed to color a given graph $G$ such that two adjacent vertices cannot share the same color. $G$ has 4 vertices, $v_1, \ldots, v_4$, and the following set of edges $G_E=\{(v_1,v_2), (v_1,v_3), (v_2,v_3), (v_3,v_4)\}$. Furthermore, there are 4 different colors available $c_1, \ldots, c_4$.

When encoding the problem into \texttt{pwcnf} the user could provide either the following VERTEX-based or COLOR-based partition scheme:

\begin{center}
\resizebox{0.6\columnwidth}{!}{%
\begin{tikzpicture}
    \filldraw[fill=green!20, draw=green!60] (-0.5,-0.2) ellipse (0.9cm and 1.4cm);
    \filldraw[fill=yellow!20, draw=yellow!60] (1.5,-0.2) ellipse (0.9cm and 1.4cm);
    \filldraw[fill=orange!20, draw=orange!60] (3.5,-0.2) ellipse (0.9cm and 1.4cm);
    \filldraw[fill=red!20, draw=red!60] (5.5,-0.2) ellipse (0.9cm and 1.4cm);
    
    \filldraw[fill=green!20, draw=green!60] (-0.5,-4.6) ellipse (0.9cm and 1.4cm);
    \filldraw[fill=yellow!20, draw=yellow!60] (1.5,-4.6) ellipse (0.9cm and 1.4cm);
    \filldraw[fill=orange!20, draw=orange!60] (3.5,-4.6) ellipse (0.9cm and 1.4cm);
    \filldraw[fill=red!20, draw=red!60] (5.5,-4.6) ellipse (0.9cm and 1.4cm);
    
    \node at (2.5,2.2) {\emph{VERTEX-based}};
    \node at (-0.5,1.5) {$V_1$};
    \node at (1.5,1.5) {$V_2$};
    \node at (3.5,1.5) {$V_3$};
    \node at (5.5,1.5) {$V_4$};
    
    \node at (2.5,-2.2) {\emph{COLOR-based}};
    \node at (-0.5,-2.9) {$C_1$};
    \node at (1.5, -2.9) {$C_2$};
    \node at (3.5,-2.9) {$C_3$};
    \node at (5.5,-2.9) {$C_4$};

    \node at (-0.5,0.7) {$\neg X_{v_1}^{c_1}$};
    \node at (-0.5,0.1) {$\neg X_{v_1}^{c_2}$};
    \node at (-0.5,-0.5) {$\neg X_{v_1}^{c_3}$};
    \node at (-0.5,-1.1) {$\neg X_{v_1}^{c_4}$};
    \node at (1.5,0.7) {$\neg X_{v_2}^{c_1}$};
    \node at (1.5,0.1) {$\neg X_{v_2}^{c_2}$};
    \node at (1.5,-0.5) {$\neg X_{v_2}^{c_3}$};
    \node at (1.5,-1.1) {$\neg X_{v_2}^{c_4}$};
    \node at (3.5,0.7) {$\neg X_{v_3}^{c_1}$};
    \node at (3.5,0.1) {$\neg X_{v_3}^{c_2}$};
    \node at (3.5,-0.5) {$\neg X_{v_3}^{c_3}$};
    \node at (3.5,-1.1) {$\neg X_{v_3}^{c_4}$};
    \node at (5.5,0.7) {$\neg X_{v_4}^{c_1}$};
    \node at (5.5,0.1) {$\neg X_{v_4}^{c_2}$};
    \node at (5.5,-0.5) {$\neg X_{v_4}^{c_3}$};
    \node at (5.5,-1.1) {$\neg X_{v_4}^{c_4}$};
    
    \node at (-0.5,-3.7) {$\neg X_{v_1}^{c_1}$};
    \node at (-0.5,-4.3) {$\neg X_{v_2}^{c_1}$};
    \node at (-0.5,-4.9) {$\neg X_{v_3}^{c_1}$};
    \node at (-0.5,-5.5) {$\neg X_{v_4}^{c_1}$};
    \node at (1.5,-3.7) {$\neg X_{v_1}^{c_2}$};
    \node at (1.5,-4.3) {$\neg X_{v_2}^{c_2}$};
    \node at (1.5,-4.9) {$\neg X_{v_3}^{c_2}$};
    \node at (1.5,-5.5) {$\neg X_{v_4}^{c_2}$};
    \node at (3.5,-3.7) {$\neg X_{v_1}^{c_3}$};
    \node at (3.5,-4.3) {$\neg X_{v_2}^{c_3}$};
    \node at (3.5,-4.9) {$\neg X_{v_3}^{c_3}$};
    \node at (3.5,-5.5) {$\neg X_{v_4}^{c_3}$};
    \node at (5.5,-3.7) {$\neg X_{v_1}^{c_4}$};
    \node at (5.5,-4.3) {$\neg X_{v_2}^{c_4}$};
    \node at (5.5,-4.9) {$\neg X_{v_3}^{c_4}$};
    \node at (5.5,-5.5) {$\neg X_{v_4}^{c_4}$};

\end{tikzpicture}
}
\vspace*{-3mm}
\end{center}
\end{example}

\section{Seating Assignment MaxSAT encoding}
\label{sec:assignment:encoding}

In this section, we present the MaxSAT encoding used for the seating assignment where the goal is to seat persons at tables such that the following properties hold:
\begin{itemize}
    \item Each table has a minimum and maximum number of persons;
    \item Each person is seated at exactly one table;
    \item Each person has some tags that represent their interests; if a person $p$ is seated at a table, then that table has all the tags of $p$.
    \item The optimization goal is to minimize the number of different tags that appear at each table.
\end{itemize}

Consider a seating problem with $p$ persons and $t$ tables, where each table has at least $l$ persons and at most $m$ persons. Assume that the set of tags is defined by $G$ and the set of tags each person $p$ has is defined by $p(G)$. Assume also that the set of tables is defined by $T$ and the set of persons is defined by $P$. %

\subsection*{Variables.} We model the problem by creating $Y$ variables that represent if a given table has a tag and $X$ variables that assign persons to tables. The Boolean variables $Y_t^g$ with $t\in T$ and $g \in G$ are assigned to 1 if there is \emph{at least one person} $p$ with a tag $g$ that is seated at table $t$. The Boolean variables $X_t^p$ with $t\in T$ and $p \in P$ are assigned to 1 if person $p$ is seated at table $t$.

\subsection*{Optimization criteria (unweighted).} The goal of this optimization problem is to minimize $\sum_{{t\in T},{g \in G}} Y_t^g$ subject to the constraints of the problem. Note that minimizing this sum is equivalent to create unit soft clauses $\neg Y_t^g$.

\subsection*{Constraints.} For simplicity, we will represent constraints for at-least-$k$, at-most-$k$, and exactly-$k$ using linear equations. However, these constraints can be encoded into CNF by using a cardinality encoding such the Sinz encoding~\cite{sinz-cp05}. We use the $X$ and $Y$ variables to encode the following hard constraints.

\begin{itemize}
    \item Each table has at most $m$ persons:
    \begin{itemize}
        \item ${\bigforall_{t \in T}} \sum_{p\in P} X_t^p \leq m$
    \end{itemize}
    
    \item Each table has at least $l$ persons:
    \begin{itemize}
        \item ${\bigforall_{t \in T}} \sum_{p\in P} X_t^p \geq l$
    \end{itemize}
    
    \item Each person is seated at exactly one table:
    \begin{itemize}
        \item ${\bigforall_{p \in P}} \sum_{t\in T} X_t^p = 1$
    \end{itemize}
    
    \item If a person is seated at table $t$ then that table has all the tags of $p$:
    \begin{itemize}
        \item $\bigforall_{t \in T} \bigforall_{p \in P} \bigforall_{g \in p(G)} \neg X^p_t \vee Y^g_t$
    \end{itemize}
\end{itemize}

When encoding the problem into MaxSAT, the $\neg Y_t^g$ literals will correspond to unit soft clauses. The user can potentially group these soft clauses in two distinct ways: (1) variables that share the same tag are grouped, or (2) variables that share the same table are grouped. We call the former partition scheme \emph{TAGS-based} and the latter \emph{TABLES-based}. 

\begin{example}
Consider that Alice wants to seat 5 persons, $p_1, \ldots, p_5$, in two tables $t_1, t_2$. Each table must have at least 2 persons and at most 3 persons. Each person has a set of interests described by their tags as follows: 
\[
p_1 = \{A, B\}, p_2 = \{C\},p_3 = \{B\}, p_4 = \{C,A\}, p_5 = \{A\}
\]

When encoding the problem into \texttt{pwcnf} the user could provide either the following TAGS-based or TABLES-based partition scheme:

\begin{center}
\vspace*{-3mm}
\begin{tikzpicture}
    \filldraw[fill=green!20, draw=green!60] (-1.5,0) ellipse (0.8cm and 1.2cm);
    \filldraw[fill=yellow!20, draw=yellow!60] (0.5,0) ellipse (0.8cm and 1.2cm);
    \filldraw[fill=orange!20, draw=orange!60] (2.5,0) ellipse (0.8cm and 1.2cm);
    
    \filldraw[fill=green!20, draw=green!60] (5.5,0) ellipse (0.8cm and 1.2cm);
    \filldraw[fill=yellow!20, draw=yellow!60] (7.5,0) ellipse (0.8cm and 1.2cm);
    
    \node at (0.5,2.2) {\emph{TAGS-based}};
    \node at (-1.5,1.5) {$A$};
    \node at (0.5,1.5) {$B$};
    \node at (2.5,1.5) {$C$};
    
    \node (x1) at (4,2.5) {};
    \node (x2) at (4,-1.5) {};
    \draw[thick] (x1) -- (x2);
    
    \node at (6.5,2.2) {\emph{TABLES-based}};
    \node at (5.5,1.5) {$t_1$};
    \node at (7.5,1.5) {$t_2$};

    \node at (-1.5,0.5) {$\neg Y_{t_1}^A$};
    \node at (-1.5,-0.3) {$\neg Y_{t_2}^A$};
    \node at (0.5,0.5) {$\neg Y_{t_1}^B$};
    \node at (0.5,-0.3) {$\neg Y_{t_2}^B$};
    \node at (2.5,0.5) {$\neg Y_{t_1}^C$};
    \node at (2.5,-0.3) {$\neg Y_{t_2}^C$};
    
    \node at (5.5,0.7) {$\neg Y_{t_1}^A$};
    \node at (5.5,0) {$\neg Y_{t_1}^B$};
    \node at (5.5,-0.7) {$\neg Y_{t_1}^C$};
    \node at (7.5,0.7) {$\neg Y_{t_2}^A$};
    \node at (7.5,0) {$\neg Y_{t_2}^B$};
    \node at (7.5,-0.7) {$\neg Y_{t_2}^C$};
\end{tikzpicture}
\vspace*{-3mm}
\end{center}
\end{example}

\section{Cactus Plots}
\label{appx:plots}

\begin{figure}[t!]
    \centering
    \resizebox{0.75\columnwidth}{!}{%
    \includegraphics{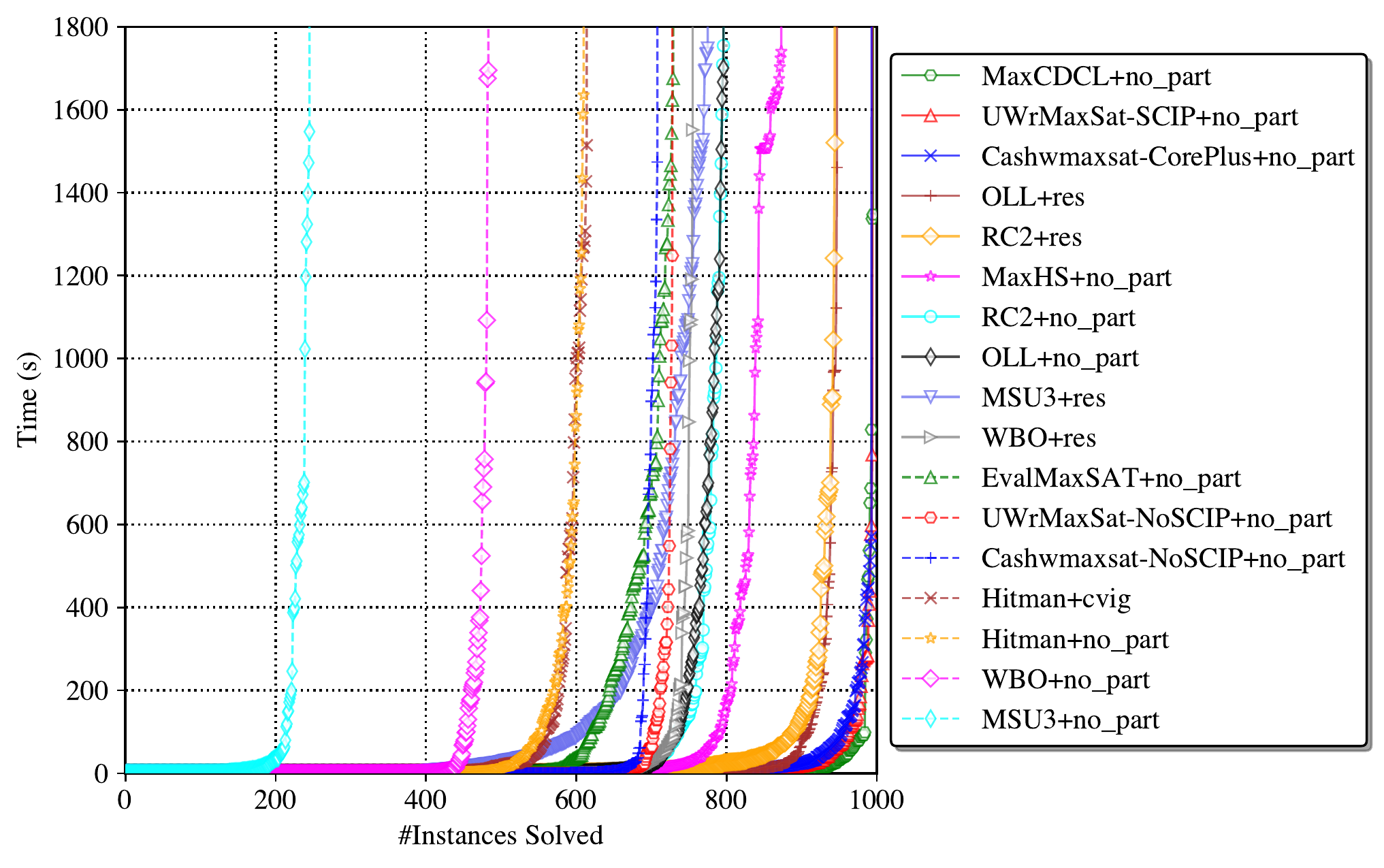}
    }
    \caption{Cactus plot - The number of instances solved for the MSC problem for each MaxSAT algorithm and partitioning scheme.}
    \label{fig:cactus-msc}
\end{figure}

Figure~\ref{fig:cactus-msc} shows the cactus plot for the set
of instances solved for the MSC problem for each MaxSAT algorithm and the partitioning scheme used. 
The plot presents the solving time ($y$-axis) against the number of 
solved instances ($x$-axis).  The legend in Figure~\ref{fig:cactus-msc} is 
sorted in decreasing order of number of solved instances. 
This plot supports the  results presented in Table~\ref{tab:msc}.
One can clearly see a gap between some algorithms using some partitioning scheme and not using any partitioning scheme (e.g., MSU3-RES/MSU3-NoPart, OLL-RES/OLL-NoPart).

\begin{figure}[t!]
    \centering
    \resizebox{0.75\columnwidth}{!}{%
    \includegraphics{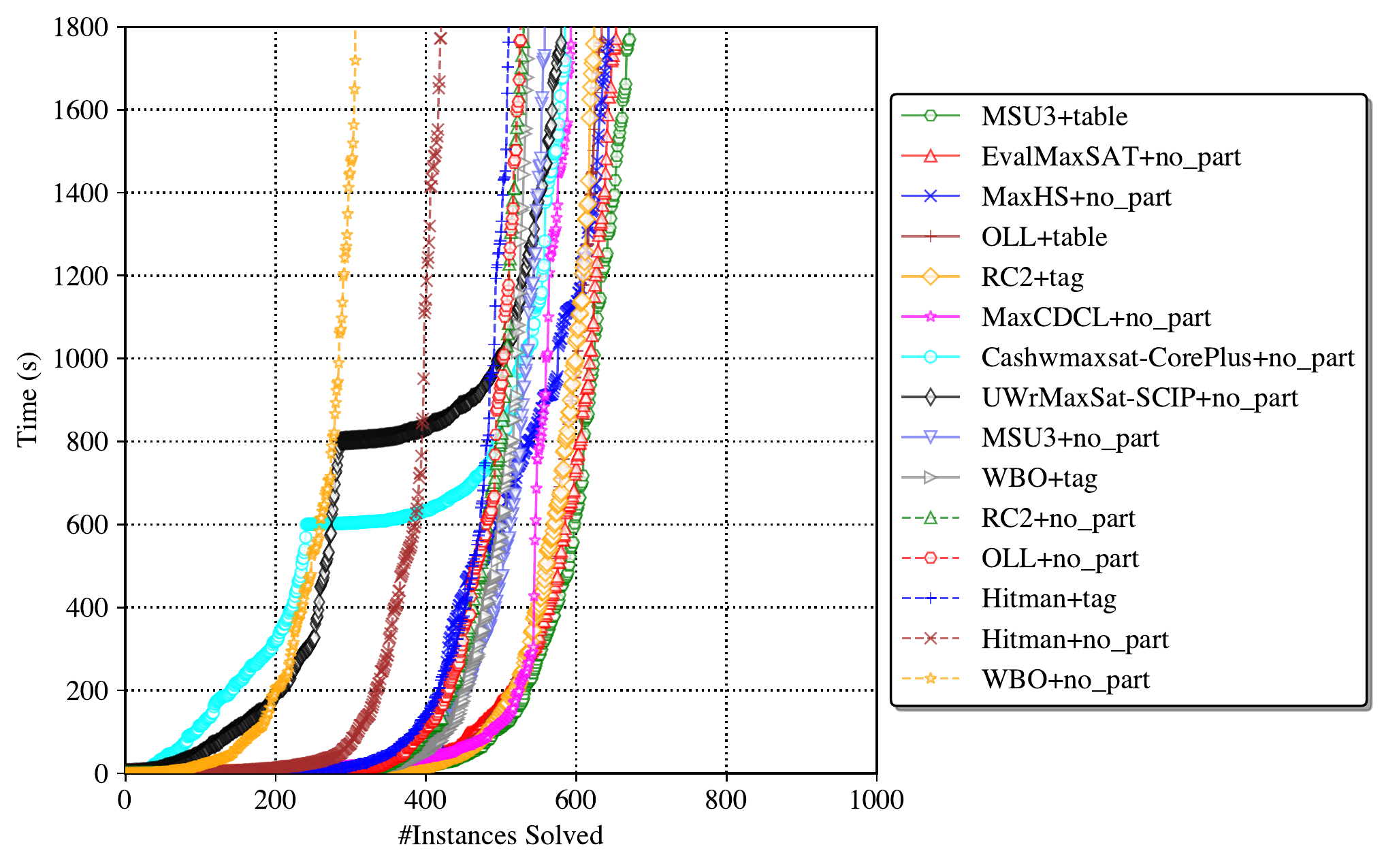}
    }
    \caption{Cactus plot - The number of instances solved for the Seating Assignment problem for each MaxSAT algorithm and partitioning scheme.}
    \label{fig:cactus-seating}
\end{figure}

Figure~\ref{fig:cactus-seating} shows the cactus plot for the set
of instances solved for the Seating Assignment problem for each MaxSAT algorithm and the partitioning scheme used. The legend in Figure~\ref{fig:cactus-seating} is also sorted in decreasing order of the number of solved instances. 
This plot supports the  results presented in Table~\ref{tab:seating}. One can see that, on this set of instances, using some partitioning scheme there is an improvement in the performance of the algorithms available on \upmax, i.e., MSU3, OLL, WBO, Hitman, RC2.

\end{document}